\documentclass[acmsmall,screen,nonacm]{acmart}

\usepackage{amsfonts}
\usepackage{booktabs}
\usepackage{tabularx}
\usepackage{siunitx}
\usepackage{graphicx}
\usepackage{subcaption}
\usepackage{array}
\usepackage{makecell} 
\usepackage{subcaption}
\usepackage{fancyhdr}
\usepackage[ruled,vlined]{algorithm2e} 

\AtBeginDocument{%
  }

\setcopyright{none}
\renewcommand\footnotetextcopyrightpermission[1]{}
\renewcommand{\authorsaddresses}{}

\makeatletter
\def\@mkauthorsaddresses{}
\makeatother

\begin{document}

\title{Thread-Efficient Decoding for Neural Texture Compression}

\author{Janarbek Matai}
\affiliation{%
  \institution{Advanced Micro Devices, Inc.}
  \country{USA}
}

\author{Sho Ikeda}
\affiliation{%
  \institution{Advanced Micro Devices, Inc.}
  \country{Japan}
}

\author{Lukasz Lipski}
\affiliation{%
  \institution{Advanced Micro Devices, Inc.}
  \country{Poland}
}

\author{Takahiro Harada}
\affiliation{%
  \institution{Advanced Micro Devices, Inc.}
  \country{USA}
}

\renewcommand{\shortauthors}{Matai et al.}

\begin{abstract}
Neural texture compression (NTC) achieves higher compression ratios than BCn formats but suffers from GPU thread divergence, which significantly reduces runtime performance. 
In this work, we propose a shared decoder MLP architecture — trained with a gradual decoder freezing schedule — combined with texture clustering to reduce thread divergence by 25\%–52\% while preserving rendering quality
We evaluate our method on over 500 textures and multiple real rendering scenes, demonstrating up to 8.48× speedup on the Radeon RX 9070 XT GPU compared to non-shared baselines. Our key contributions include: (1) a unified shared decoder architecture that reduces divergence by grouping textures; (2) a training recipe with gradual decoder freezing that improves stability and reconstruction accuracy; (3) a semantic clustering strategy using CLIP embeddings that groups similar textures for effective decoder sharing; and (4) comprehensive performance and ablation studies validating our approach.


\end{abstract}  

\maketitle  
\renewcommand{\authorsaddresses}{}

\begin{figure*}
\centering
\begin{minipage}{0.32\linewidth}
    \centering
    \includegraphics[width=\linewidth]{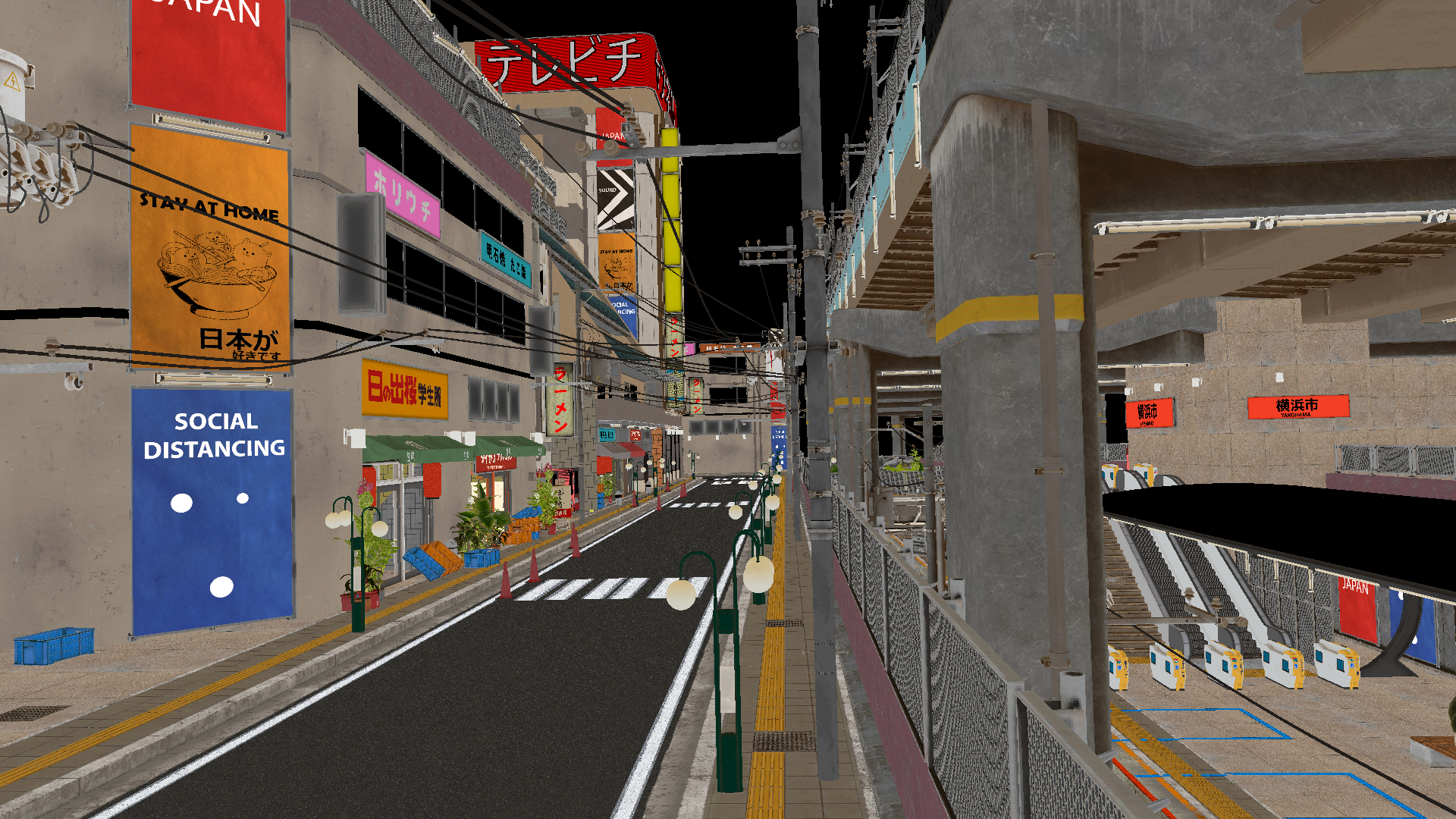}
    \\(d)
\end{minipage}\hfill
\begin{minipage}{0.32\linewidth}
    \centering
    \includegraphics[width=\linewidth]{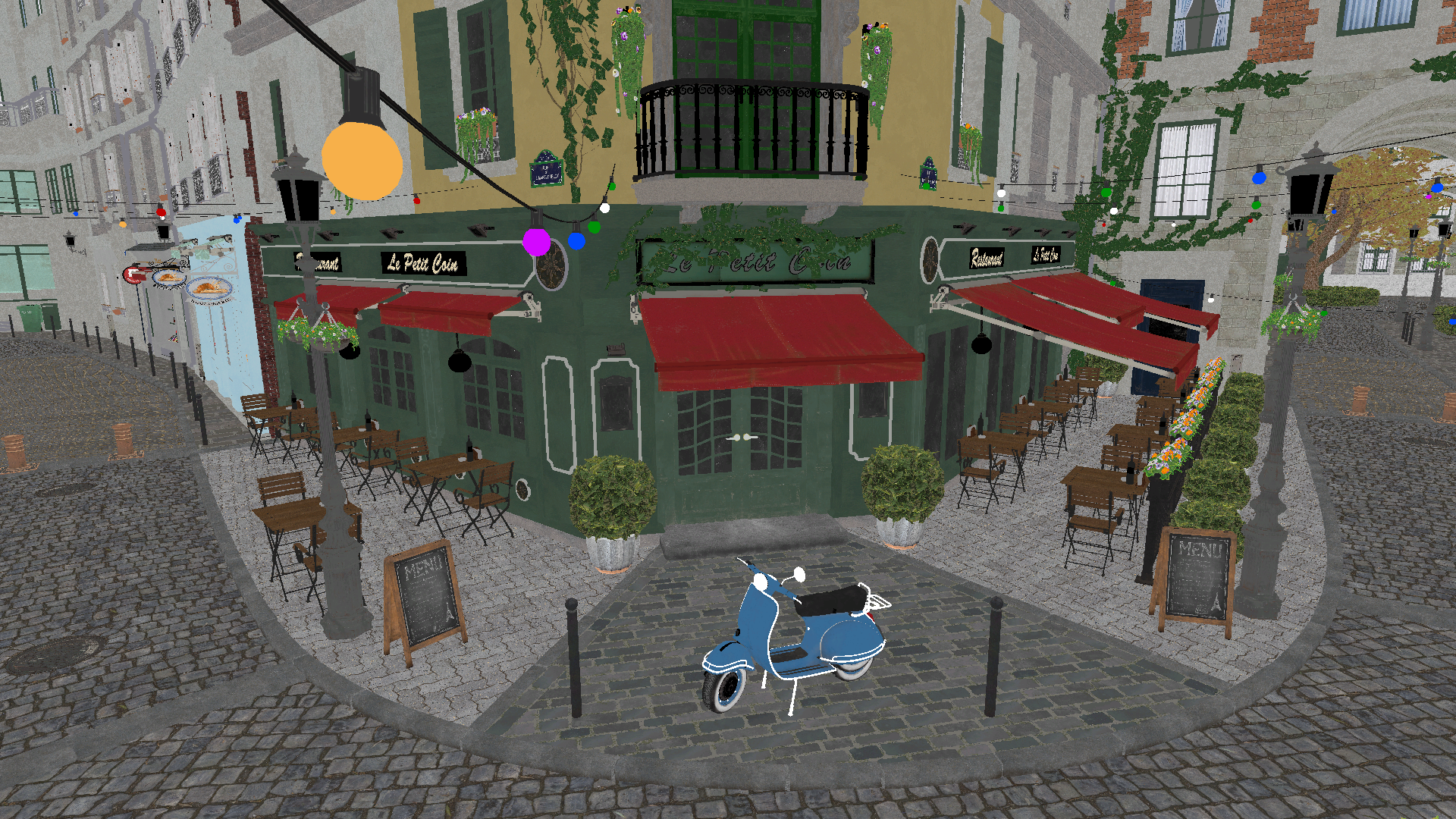}
    \\(e)
\end{minipage}\hfill
\begin{minipage}{0.32\linewidth}
    \centering
    \includegraphics[width=\linewidth]{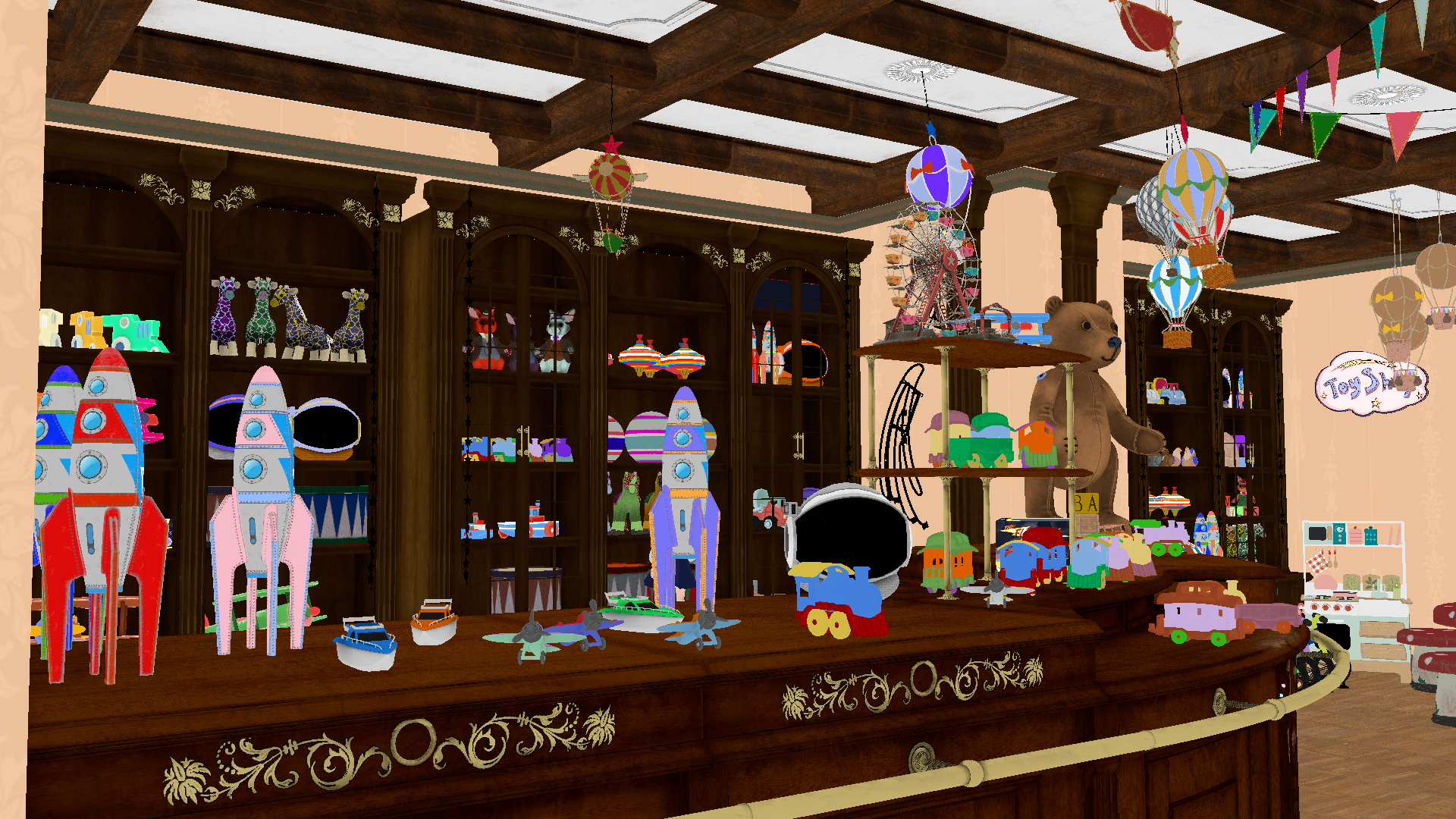}
    \\(f)
\end{minipage}

\vspace{0.5em} 

\begin{minipage}{0.32\linewidth}
    \centering
    \includegraphics[width=\linewidth]{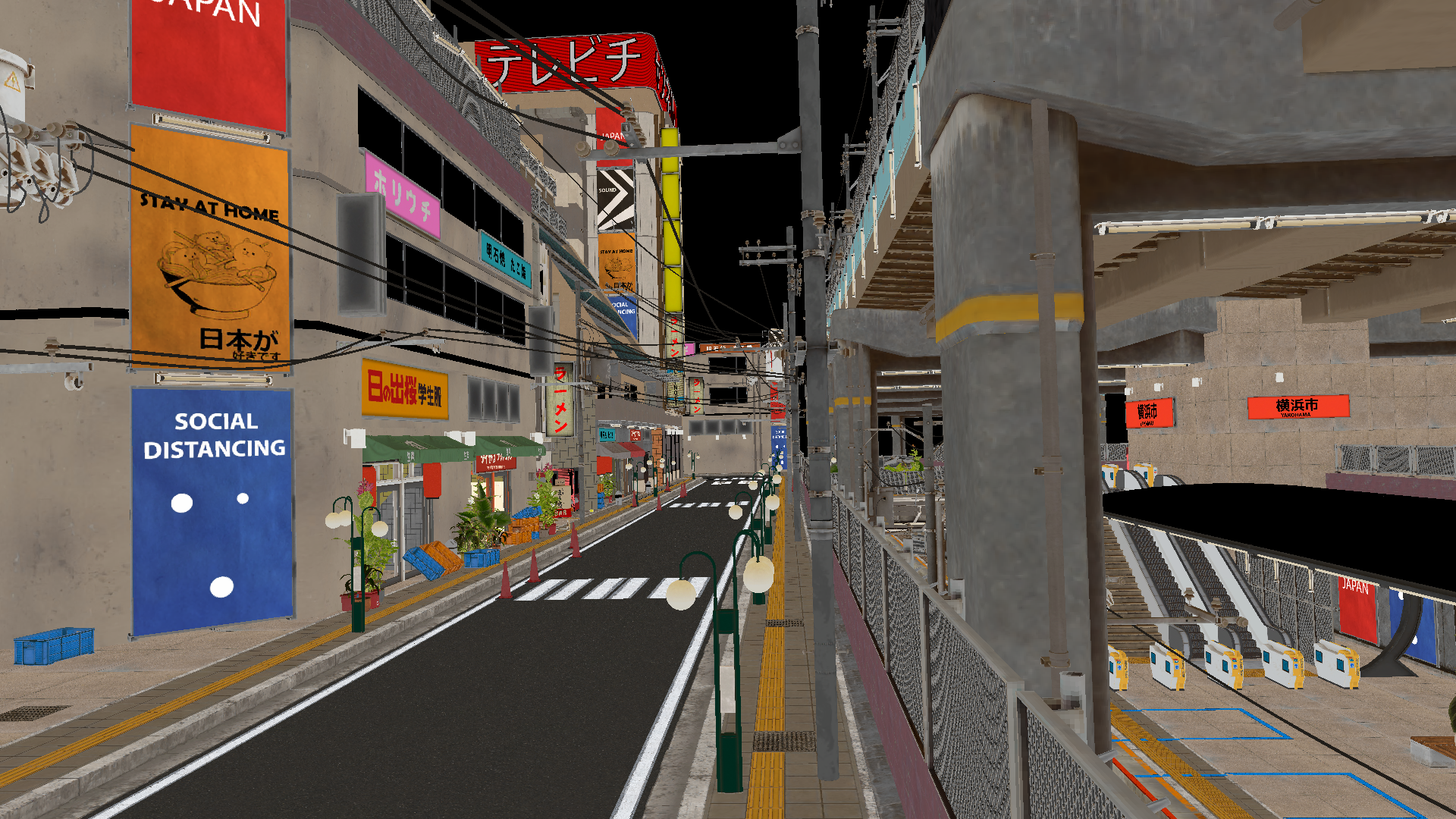}
    \\(a)
\end{minipage}\hfill
\begin{minipage}{0.32\linewidth}
    \centering
    \includegraphics[width=\linewidth]{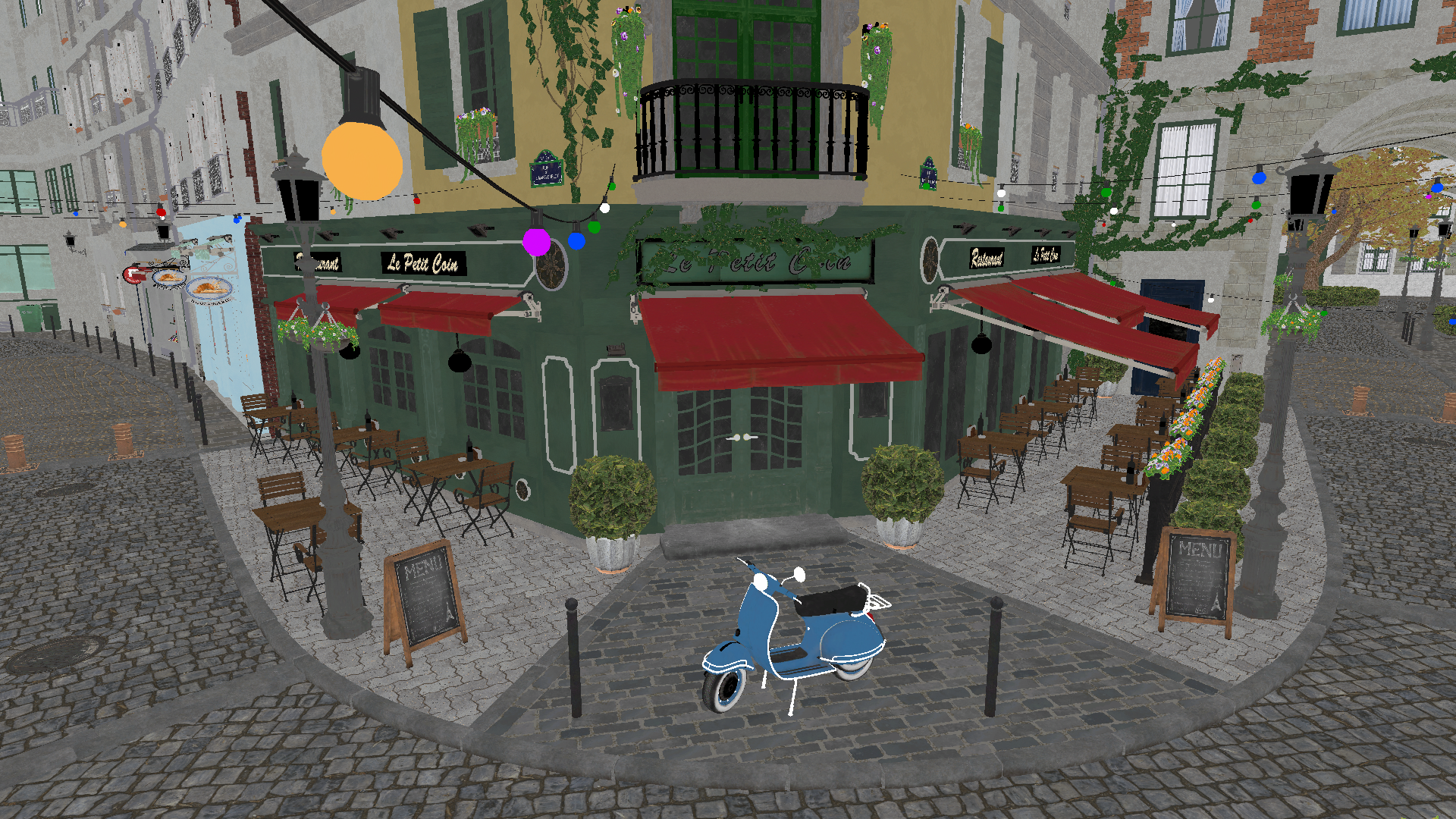}
    \\(b)
\end{minipage}\hfill
\begin{minipage}{0.32\linewidth}
    \centering
    \includegraphics[width=\linewidth]{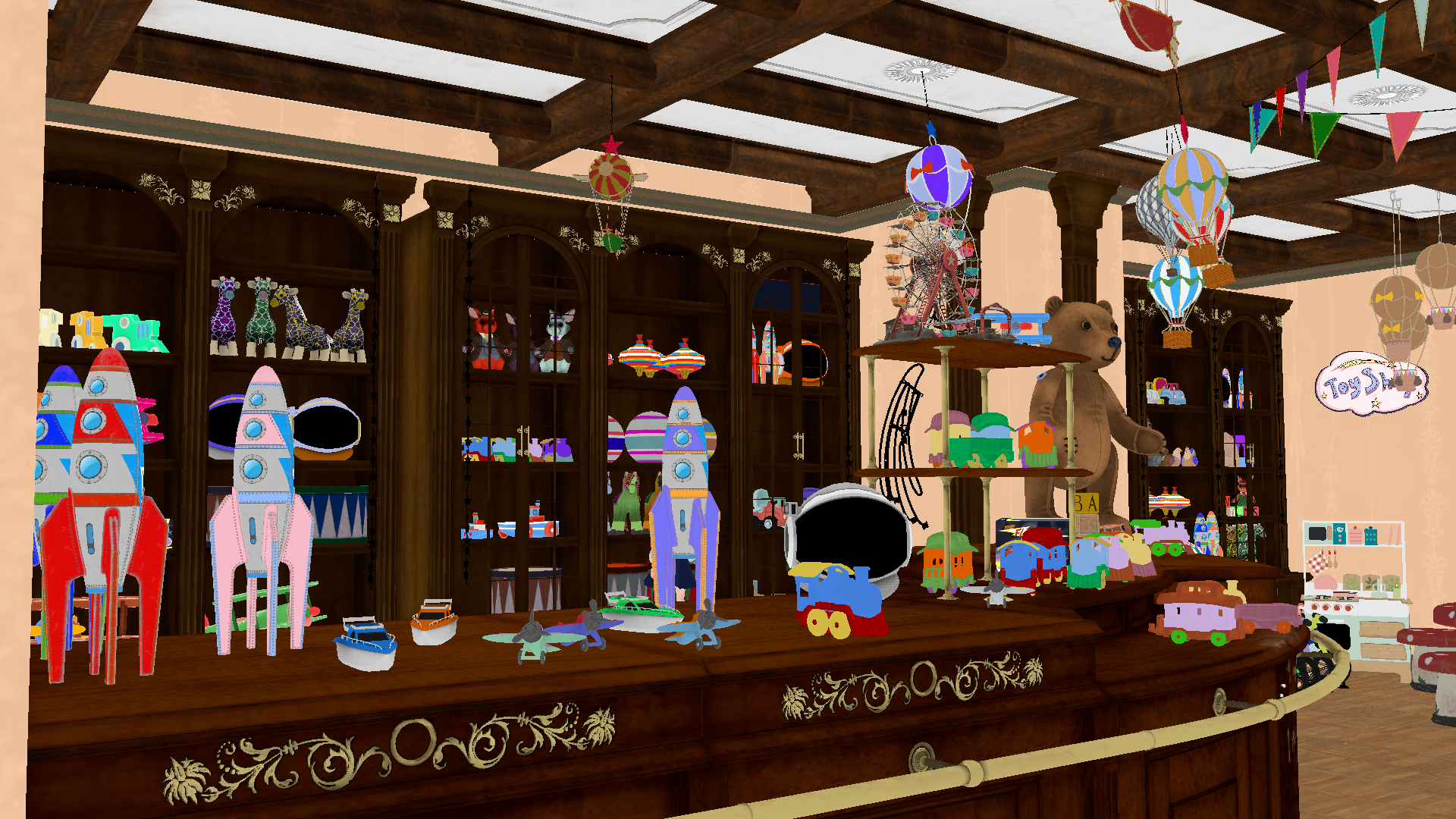}
    \\(c)
\end{minipage}

\vspace{0.5em} 


\begin{minipage}{1\linewidth}
    \centering
    \includegraphics[width=\linewidth]{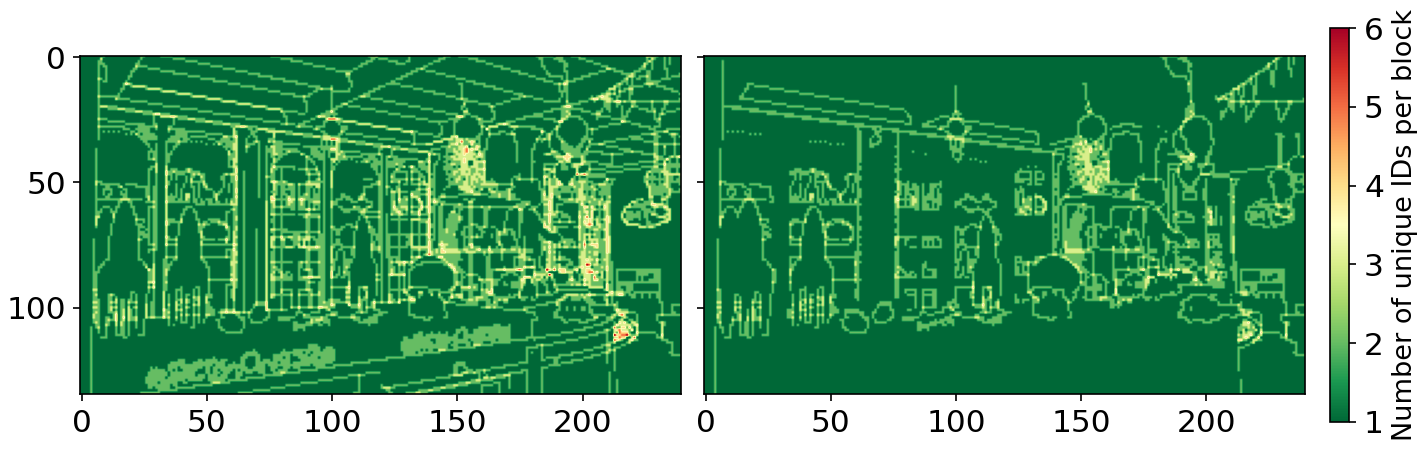}
    \\(h)
\end{minipage}

\caption{Top row: Rendered original scenes for Yokohama, Bistro, and ToyShop using the default textures, middle row: Rendered scenes for Yokohama, Bistro, and ToyShop using the textures generated by Clustered network, bottom row: Heatmap of divergence maps for ToyShop scene. The right side of each graph shows the thread divergence in Baseline network, and left side shows how much improved we are getting to reduce the thread divergence using Clustered--green color indicates less divergence. ToyShop have the highest 6 on divergence values as indicated by colorbar.}
\label{fig:real-scenes-rendered}
\end{figure*}

\section{Introduction}

Texture quality plays a crucial role in game graphics by adding detail that cannot be represented purely by geometry. As display resolutions have increased, demands on texture quality have grown accordingly, leading to a large memory and storage footprint for texture data. Neural texture compression (NTC) uses neural networks to compress textures and has been shown to achieve higher compression ratios than traditional analytical methods ~\cite{nvidNTC23, fujieda2024neural, ubi24, farhadzadeh2024neural}.

Typical NTC models consist of two parts: an encoder for a latent representation of the texture (e.g., latent pyramids or grids) and a dedicated decoder implemented as a multi-layer perceptron (MLP). While the resulting size reduction is highly beneficial, NTC introduces new challenges for runtime implementation due to the need to query different MLP weights (tensors) during rendering~\cite{laurent2025hardware}. Tensor operations, unlike traditional GPU arithmetic, must be executed across an entire wave rather than per lane in the GPU’s SIMD unit~\cite{amd_gpuopen_wmma_rdna3}. Modern GPUs are optimized for cases where a single tensor operation or neural network is processed per wave; efficiency drops when many networks must be evaluated within the same wave due to \textit{thread divergence}. \textit{This divergence can occur frequently when neural textures are evaluated in the middle of the rendering pipeline}. Designing NTC inference that handles divergent threads efficiently, while remaining easy for developers to use, remains an open problem.

In this work, we propose using \textit{a shared decoder} for a groups of textures in a rendering scene. The decoder is typically implemented as a multi-layer perceptron (MLP) in texture compression; we use the term “decoder” throughout the paper. Instead of using one decoder per texture set or texture, as in existing NTC methods, we use a single decoder for multiple texture sets. The extreme case is a single decoder for all textures in the scene, which completely eliminates thread divergence but reduces rendering quality, as we show later. The shared decoder approach proposed in this paper, shifts the burden from the decoder to individual encoders. In this paper, we provide a NTC model architecture with \textit{a training strategy} that reduces the thread divergence significantly, and we provide theoretical and empirical performance analysis that demonstrate that our method speeds up the texture decompression by almost $5X-8X$. We also provided empirical analysis how much it reduces the thread divergence in diverse scenarios during rendering. 


The main contributions of this paper are as follows:
\begin{itemize}
   \item Unified (Shared) decoder ML architecture and its impact on reducing thread divergence
   \item Training recipe for the shared decoder NTC model that provides stability to training which in turn resulted better accuracy and stability when share the decoder and critical for success of our method.
   \item Comprehensive study of the shared decoder for more than 500 textures and real scenes, and we provided empirical study of how our method reduces thread divergence for three real scenes reconstructions. To our best of knowledge, all existing NTC work used small number of textures (usually less than 20), and we are the first to study more than 600 textures.  
   \item  We further introduce neural network architectures that reduce the number of decoders in the scene by clustering textures into different groups. We evaluate several clustering strategies (baseline, clustered, random) and show that machine learning–based clustering achieves the best results compared to the other approaches. 
    \item Finally, we provide performance analysis of our method that demonstrated significant speed up, and also provided ablation study of impact of how to train shared decoder and different network sizes. 
\end{itemize}

Figure~\ref{fig:real-scenes-rendered} shows visualizations of the rendered scenes using both the original textures (default) and the textures learned based on our method (e.g., shared decoder with clustered). The top row shows scenes rendered with the original textures—these are the default textures and are not generated using neural texture compression. The middle row shows scenes rendered with textures generated by our shared decoder neural network (in this case: Clustered). The bottom row shows the visual heatmap of how much thread divergence we can reduce from baseline vs. our proposed method in this paper. As the figure illustrates, our method produces rendered scenes whose visual quality is almost indistinguishable from that of the default textures while having less thread divergence or almost $5X-8X$ faster texture compression.

The primary advantage of neural texture compression is the reduced memory footprint of textures; however, it increases runtime cost, particularly due to thread divergence. These results demonstrate that we can reduce thread divergence by sharing the decoder while still rendering scenes with nearly the same visual quality as the default textures.

Existing approach to reducing thread divergence focus on cooperative vectors (low-level optimization)~\cite{laurent2025hardware}. All prior NTC work, however, trains neural networks on a per-texture or per-texture-set basis and does not explicitly address thread divergence. \textit{To the best of our knowledge, this is the first work} to use a shared decoder approach  to tackle thread divergence. Moreover, we provide comprehensive study how we share decoder among textures and studied different clustering methods such as machine-learning-based clustering. Our method is orthogonal and applicable to all existing NTC approaches, providing a clear trade-off between accuracy and reduced thread divergence. In the next sections, we define thread divergence, present our approach, and evaluate its benefits on both existing texture datasets and real scenes, and finally we present performance analysis and ablation study. 

\begin{figure}
    \centering
    \includegraphics[width=0.8\linewidth]{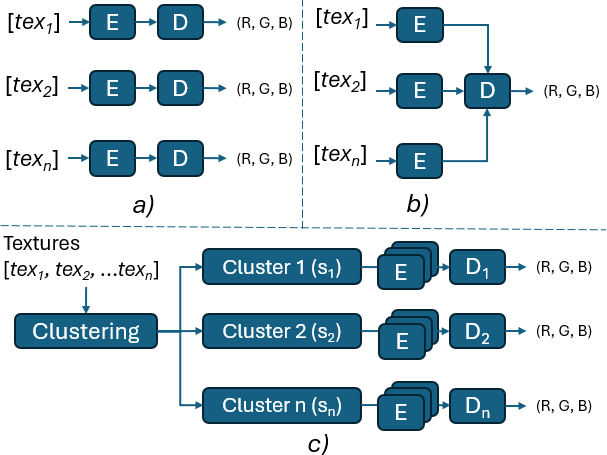}
    \caption{(a) Existing NTC architecture. (b) Shared decoder. (c) Similar textures share a decoder based on clustering where $s_i$ is size of cluster $i$ as well as number of encoders, and there are $n$ clusters}
    \label{fig:method_overview}
\end{figure}




\section{Thread Divergence in NTC}
\label{sec:thread_divergence}
We aim to reduce thread divergence in NTC. Thread (or wave) divergence is a GPU bottleneck that occurs when threads in the same wave take different execution paths. Modern NTCs use an encoder–decoder architecture, where the decoder consists of several MLP layers. In NTC rendering, divergence arises when different pixel locations are decoded by MLPs using different textures. To measure the benefits of our method, we introduce a thread divergence map, or simply divergence map as follows: 

The \emph{divergence map} \(D \in \mathbb{N}^{H_b \times W_b}\) for a given scene of size \(HxW\) is then defined as the number of distinct textures (we assume each texture has ID) per block:
$
D(i,j) = \bigl|\mathcal{U}_{i,j}\bigr|
= \left|
\left\{
I(y,x) \,\middle|\,
(y,x) \in \mathcal{B}_{i,j}
\right\}
\right|
$
where we assume we divide the scene into blocks, and $\mathcal{U}_{i,j}$ is the set of unique texture IDs in a block $\mathcal{B}_{i,j}$, \(I(y,x)\) indicates the
texture ID used at pixel \((y,x)\).
Intuitively, \(D(i,j) = 1\) means that all pixels in block \((i,j)\)
use the same texture (low divergence), while larger values of
\(D(i,j)\) indicate that many different textures are accessed within
the same block (high divergence).








\textbf{Quantifying thread divergence:} 
Let $X \in \mathbb{R}^{n \times m}$ be a divergence map whose ideal configuration consists of all entries equal to one. 
We measure the deviation of $X$ from this ideal by computing the average absolute difference between each entry and one:
\begin{equation}
\label{eq:normalized_ones_metric}
\mathcal{D}(X) = \frac{1}{nm} \sum_{i=1}^{n} \sum_{j=1}^{m} \left| X_{ij} - 1 \right|.
\end{equation}
This metric equals zero if and only if $X$ is an all-ones matrix (no thread divergence), and increases with both the magnitude and variability of the entries, while remaining invariant to matrix size.
If all threads in a block uses the same textures, then we have a divergence score of 0. In next section, we propose a solution to reduce thread divergence, and in Section~\ref{sec:results}, we provide empirical results that our method improves the thread divergence including a visualization of divergence map as a heatmap for real scenes.

\section{Methodology}


\subsection{Shared Decoder}
To reduce the divergence of neural texture execution, a solution would be using the same decoder for all the textures. However, as we are going to show later, we observed quality degradation from this approach. Thus we propose to introduce clustering of textures which results in smaller number of decoders than baseline where one decoder is trained for each texture set, but larger number of decoders than using single decoder. Introducing a shared decoder, particularly with clustering, reduces thread divergence while improving quality compared to using a single decoder, as we show in the results section
Figure~\ref{fig:method_overview} provides an overview of existing and our proposed methods. The fundamental difference between our method and the state of the art is that existing NTC methods train the encoder and decoder on a per-texture or per-texture-set basis, as shown in Figure~\ref{fig:method_overview} (a). In our work, we propose two fundamental changes. First, we share the decoder, as shown in Figure~\ref{fig:method_overview} (b), to reduce thread divergence in the decoder. Second, we build on (b) and add clustering of textures in our method. The idea is that similar textures will be grouped together and share the same decoder, as shown in Figure~\ref{fig:method_overview} (c). In both, b) and c) each texture(s) have their own encoder as a), but with shared decoder. For clustering, we explored different algorithms, such as machine learning-based clustering and random grouping of textures, which we outline in the results section. In general, clustering groups high-level, semantically related textures (e.g., wood textures, metal surfaces, fabric patterns) that likely share similar statistical properties and visual features. We believe that a decoder trained on similar textures can specialize in those characteristics, leading to better reconstruction quality than using a single decoder, as in Figure~\ref{fig:method_overview}(b).


\begin{algorithm}[H]
\DontPrintSemicolon
\SetAlgoLined
\SetKwFunction{SelfFreeze}{SelfFreezeDecoderGradually}
\SetKwProg{Fn}{Function}{:}{}
\caption{Gradual Decoder Freezing Scale}
\label{alg:self_freeze_decoder_gradually}

\Fn{\SelfFreeze{epoch, freeze\_start\_epoch, freeze\_duration}}{
    \If{freeze\_start\_epoch is None \textbf{ or } epoch $<$ freeze\_start\_epoch}{
        \Return $1.0$\;
    }
    \ElseIf{epoch $\ge$ freeze\_start\_epoch + freeze\_duration}{
        \Return $0.0$ \tcp*{fully frozen}
    }
    \Else{
        $progress \gets \dfrac{epoch - \text{freeze\_start\_epoch}}{\text{freeze\_duration}}$\;
        \Return $0.5 \cdot \left(1 + \cos(\pi \cdot progress)\right)$\;
    }
}
\end{algorithm}

\subsection{Training Shared Decoder}
One of the key training techniques we use to make the shared decoder work is freezing the decoder after a certain number of epochs (typically after 50\% or 75\% of the total epochs, as found empirically). 
Previous studies have investigated \textit{freezing parts of large neural networks}, such as encoder or decoder components, for various tasks including task and domain adaptation or transfer learning \cite{thompson2018freezing, stickland2021recipes, dhole2025multi, son2024not, song2024rethinking} for adaptation of network to new data, stabilizing training. In contrast, we study for the first time \textit{freezing the decoder of a small encoder--decoder network of NTC}, which poses additional challenges for adaptation, and we demonstrate strategies to overcome these limitations.
In this work, freezing the decoder helps to adapt the individual encoders to individual textures, and help the decoder generalize well. 
To improve the stability and final accuracy of the encoder--decoder network, we do not freeze the decoder parameters abruptly. Instead, we apply a \emph{gradual decoder freezing} schedule that smoothly reduces the decoder's effective learning contribution over a fixed number of epochs. Let $e$ denote the current training epoch, $e_{\text{start}}$ the epoch at which the freezing process begins, and $T$ the duration (in epochs) over which the decoder is gradually frozen. We define a scalar scaling factor $\lambda(e) \in [0, 1]$ that modulates the decoder's update strength (e.g., by scaling its gradients or learning rate).
The scaling factor is calculated as 
$\lambda(e) = \frac{1}{2}\bigl(1 + \cos(\pi\, p(e))\bigr)$ where $p(e) = \operatorname{clamp}\!\left(\frac{e - e_s}{T},\, 0,\, 1\right)$.

Freezing the decoder improves both stability and accuracy. Algorithm~\ref{alg:self_freeze_decoder_gradually} describes our strategy for gradually freezing the decoder.

With our training strategy, we address two main questions regarding freezing: (1) Does freezing the decoder help? and (2) When should we freeze the decoder? Later we provide a comprehensive study across different neural network architectures and multiple texture datasets, that we demonstrate that freezing the decoder leads to higher accuracy compared to not freezing it.



\begin{algorithm}[t]
\caption{Overall texture processing pipeline}
\label{alg:pipeline}
\DontPrintSemicolon
\KwIn{
    Set of $n$ textures $\{\mathbf{X}_i\}_{i=1}^n$, 
    $\mathbf{X}_i \in \mathbb{R}^{H \times W \times 3}$; \\
    Number of groups (clusters) $m$
}
\KwOut{
    Group assignments $\{c(i)\}_{i=1}^n$; \\
    Group-wise encoder--decoder networks 
    $\big\{ \{\mathcal{E}_{g,i}\}_{i \in \mathcal{T}_g}, \mathcal{D}_g \big\}_{g=1}^m$
}

\BlankLine
\textbf{Stage 1: CLIP feature extraction}\;
\For{$i \gets 1$ \KwTo $n$}{
    $\mathbf{f}_i \gets \mathrm{CLIP}(\mathbf{X}_i)$ \tcp*{$\mathbf{f}_i \in \mathbb{R}^{512}$}
}
$\mathbf{F} \gets [\mathbf{f}_1, \dots, \mathbf{f}_n]^\top \in \mathbb{R}^{n \times 512}$\;
\tcp{$\mathbf{F}$ defines the cluster embedding space}

\BlankLine
\textbf{Stage 2: Clustering into $m$ groups}\;
Apply $k$-means clustering to $\mathbf{F}$ with $m$ clusters\;
Obtain cluster assignment for each texture:
$c(i) \in \{1, \dots, m\}$ for $i=1,\dots,n$\;
\For{$g \gets 1$ \KwTo $m$}{
    $\mathcal{T}_g \gets \{\, i \mid c(i) = g \,\}$\;
    $k_g \gets |\mathcal{T}_g|$ \tcp*{number of textures in group $g$}
}
\tcp{$\sum_{g=1}^m k_g = n$}

\BlankLine
\textbf{Stage 3: Group-wise encoder--decoder training}\;
\For{$g \gets 1$ \KwTo $m$}{
    Initialize group decoder $\mathcal{D}_g$\;
    \ForEach{$i \in \mathcal{T}_g$}{
        Initialize texture-specific encoder $\mathcal{E}_{g,i}$\;
    }
       
       \Repeat{convergence}{
        \ForEach{$i \in \mathcal{T}_g$}{
            $\mathbf{z}_{g,i} \gets \mathcal{E}_{g,i}(\mathbf{X}_i)$\;
            $\hat{\mathbf{X}}_i \gets \mathcal{D}_g(\mathbf{z}_{g,i})$\;
        }
        Update $\{\mathcal{E}_{g,i}\}_{i \in \mathcal{T}_g}$ and $\mathcal{D}_g$ 
        to minimize $\displaystyle
        \mathcal{L}_g = \sum_{i \in \mathcal{T}_g} \ell(\hat{\mathbf{X}}_i, \mathbf{X}_i)$\;
    }
}
\end{algorithm}

\subsection{Clustering Pipeline}
We provide details of overall pipeline and clustering details. We assume a collection of \(n\) input textures \(\{\mathbf{X}_i\}_{i=1}^n\), where each texture \(\mathbf{X}_i \in \mathbb{R}^{H \times W \times 3}\). Our pipeline proceeds in three stages: feature extraction, clustering, and group-wise encoder--decoder training. The overall pipeline of our work is depicted in Algorithm~\ref{alg:pipeline}

\subsubsection{CLIP feature extraction.}
For each texture \(\mathbf{X}_i\),   we extract a 512-dimensional feature vector using the CLIP image model \cite{openai_clip}:
\[
    \mathbf{f}_i = \mathrm{CLIP}(\mathbf{X}_i) \in \mathbb{R}^{512}, \quad i = 1,\dots,n.
\]
Collecting all features yields \(\mathbf{F} = [\mathbf{f}_1, \dots, \mathbf{f}_n]^\top \in \mathbb{R}^{n \times 512}\), which we refer to as the cluster embedding space.

\subsubsection{Clustering into \(m\) groups.}
We partition the textures into \(m\) groups by applying \(k\)-means to the CLIP embeddings \(\mathbf{F}\). The clustering assigns each texture \(\mathbf{X}_i\) to one of the \(m\) clusters:
\[
    c(i) \in \{1, \dots, m\}, \quad i = 1,\dots,n,
\]
where cluster \(g\) contains a subset \(\mathcal{T}_g = \{\, i \mid c(i) = g \,\}\) of textures. Let \(k_g = |\mathcal{T}_g|\) denote the number of textures in group \(g\), so that
\[
    \sum_{g=1}^m k_g = n.
\]

\subsubsection{Group-wise encoder--decoder networks.}
For each group \(g \in \{1,\dots,m\}\), we train a separate encoder--decoder network. Within group \(g\), we define an encoder \(\mathcal{E}_{g,i}\) for each texture \(\mathbf{X}_i\) in \(\mathcal{T}_g\), and a single decoder \(\mathcal{D}_g\) shared across all textures in that group:
\[
    \mathbf{z}_{g,i} = \mathcal{E}_{g,i}(\mathbf{X}_i), 
    \quad
    \hat{\mathbf{X}}_i = \mathcal{D}_g(\mathbf{z}_{g,i}),
    \quad
    i \in \mathcal{T}_g.
\]
Thus, there are \(m\) decoders in total, one per cluster, and each group defines its own set of texture-specific encoders together with the common decoder. The networks are trained to minimize a reconstruction loss over all textures in each group, for example:
\[
    \mathcal{L}_g = \sum_{i \in \mathcal{T}_g} \ell\big(\hat{\mathbf{X}}_i, \mathbf{X}_i\big),
\]

\begin{table*}[ht]
    \centering
    \caption{Number of textures per category.}
    \label{tab:texture_counts}
    \resizebox{\textwidth}{!}{%
        \begin{tabular}{l rrrrrrrrrrr}
            \toprule
            Category  &  Aerial & Brick & Concrete & Fabric & Floor & Metal & Plaster/Concrete & Rock & Roofing & Terrain & Wood\\
            \midrule
            \# of Tex. & 20 & 80 & 40 & 37 & 123 & 24 & 19 & 50 & 21 & 107 & 65\\
            \bottomrule
        \end{tabular}%
    }
\end{table*}

\section{Results}
\label{sec:results}

In this section, we present experimental results for two setups. In the first, we evaluate texture quality using a public dataset~\cite{polyhaven} (586 textures). In the second, we assess both quality and thread divergence by rendering real scenes with different neural texture compression methods. The first study provides a comprehensive accuracy evaluation, while the second demonstrates the benefit of reducing thread divergence in practical rendering scenes. We use three scenes for our rendering experiments: Yokohama~\cite{fabwebsite}, Bistro~\cite{ORCAAmazonBistro}, and ToyShop (created internally).
All experiments use the same encoder–decoder architecture. The grid size is $256 \times 256 \times 4$. The encoder output is concatenated with positional encoding and passed to an MLP decoder with 3 layers, where the first two have 64 neurons. We use LeakyReLU activations. The learning rates for the encoder and decoder are $4.0\times10^{-3}$ and $4.0\times10^{-4}$, respectively. We train for 350 epochs with a batch size of 32K and use the CLIP ViT-B/32 model for clustering. 



\begin{table}
    \centering
    \caption{Average PSNR of textures of different architectures.}
    \label{tab:polyahven_results}
    \begin{tabular}{lccccc}
        \toprule
        & Baseline & Single & Clustered & Random & Category \\
        \midrule
        PSNR & 34.07 & 32.61 & 33.18 & 33.02 & 33.01 \\
        \bottomrule
    \end{tabular}
\end{table}

\subsection{Experimental Results -- Accuracy Evaluation}
In this subsection, we provide accuracy evaluation for 600 textures from  Polyhaven \cite{polyhaven} dataset, next we evaluate accuracy of our method on actual rendering scenes. 

\textbf{Experimental Results -- Polyhaven}
\label{sec:exp_polyaven_setup}
We collected 586 unique textures from \cite{polyhaven}. The texture categories and the number of samples in each category are shown in Table~\ref{tab:texture_counts}. The textures have a resolution of $1024x1024$, and there are a total of 586 textures.

Without loss of generality, we define the network architectures used in our study in terms of whether they employ a single decoder or a shared decoder. For the shared-decoder setting, we consider several ways of sharing the decoder across textures:

\begin{itemize} 
\item \textbf{Baseline}: Each texture has its own encoder and its own decoder (one encoder–decoder pair per texture). 
\item \textbf{Shared Decoder–Single (Single)}: Each texture has its own encoder, but all textures share a single common decoder. 
\item \textbf{Shared Decoder–Clustered (Clustered)}: Each texture has its own encoder, and textures are assigned to clusters (20 clusters for 586 textures); all textures within a cluster share a decoder. 
\item \textbf{Shared Decoder–Random (Random)}: Each texture has its own encoder, and textures are randomly grouped; all textures within a random group share a decoder. Note that the number of textures in this strategy have the same number of textures as Clusterd. 
\item \textbf{Shared Decoder–Category (Category)}: Each texture has its own encoder, and textures are grouped by category as defined in Table~\ref{tab:texture_counts}; all textures within a category share a single decoder (e.g., 20 aerial textures share one decoder). \end{itemize} In the remainder of this paper, we refer to these architectures as \emph{Baseline}, \emph{Single}, \emph{Clustered}, \emph{Random}, and \emph{Category}. \textit{Supplemental material provides visualization of clsutered textures. }

In Table~\ref{tab:polyahven_results}, we report results for the Baseline, Single, Clustered, Random, and Category networks, averaged over 586 textures. The Baseline achieves the highest PSNR, and the Single network the lowest. The Clustered shared-decoder architecture attains a PSNR lower than the Baseline but higher than the Random, Category, and Single variants. Overall, these results suggest that the number of shared decoders is the primary factor in reducing thread divergence, while the specific grouping strategy has also non-negligible impact. Importantly, clustering based on semantics have better PSNR than other grouping strategies as evidenced by the higher PSNR of the Clustered variant relative to the Random strategy — where both strategies are configured with the same number of textures per group to ensure a fair comparison. In this work, we use semantic grouping via CLIP embeddings; we believe that incorporating low-level feature-based clustering, such as grouping based on texture frequency content, will be an important direction for future work




\textbf{Experimental Results -- Rendering Evaluation} 
\label{sec:exp_rendering_scenes}
We use three scenes: Yokohama, Bistro, and ToyShop. We extract all textures used in these scenes—107 for Yokohama, 104 for Bistro, and 117 for ToyShop—and use 10 clusters in all experiments. Because these scenes contain textures with different resolutions, we scale all textures to $1\text{k} \times 1\text{k}$ resolution.

We first evaluate reconstructed texture quality on real rendering scenes, with results reported in Table~\ref{tbl:real_scene_psnrs}. As in the previous experiment, the Baseline achieves the highest PSNR, representing the upper bound of reconstruction quality. The Clustered architecture offers a middle ground between the Single and Random variants, consistent with the trends observed on the Polyhaven dataset. Notably, in some scenes such as Bistro, the Clustered strategy achieves a substantially higher PSNR than other grouping strategies, suggesting that the choice of clustering criterion can have a meaningful impact on reconstruction quality — particularly in scenes with semantically coherent texture content

\begin{table}
    \centering
    \begin{tabularx}{\linewidth}{lXXXX}
        \toprule
        & Baseline & Clustered & Single & Random \\
        \midrule
        Yokohama        & 41.51 & 39.86 & 39.48 & 39.81 \\
        Bistro  & 48.68 & 47.02 & 45.62 & 46.04 \\
        ToyShop        & 39.40 & 37.47 & 37.37 & 37.24 \\
        \bottomrule
    \end{tabularx}
    \caption{PSNR of three rendered images for network architectures}
    \label{tbl:real_scene_psnrs}
\end{table}
Next, we provide a qualitative and quantitative measure for the three scenes rendered using different NTC methods.We render scenes using a camera without lighting and shading, showing albedo AOV which shows the quality difference of textures better.

Table~\ref{tbl:real_scene_rendered_psnr_divergence} reports the PSNR of the rendered scenes using textures from the Baseline and Cluster networks, computed against the original rendered scene for each case. Overall, the real-scene results in Table~\ref{tbl:real_scene_psnrs} and Table~\ref{tbl:real_scene_rendered_psnr_divergence} indicate that networks with a shared decoder offer a favorable trade-off among the shared-decoder architectures, substantially reducing thread divergence while maintaining acceptable accuracy. We further discuss this trade-off between thread divergence and accuracy in the next section.




\begin{table}[t]
    \centering
    \begin{tabular}{lcccc}
        \toprule
        & Baseline & Clustered  & Divergence \\
        \midrule
        Yokohama        & 33.348 & 32.733 & 14802 / 11222 (25\%)   \\
        Bistro  & 32.712 & 32.494 & 17941 / 13373 (25\%)  \\
        ToyShop        & 34.220 & 33.687 & 13214 / 6407 (52\%) \\
        \bottomrule
    \end{tabular}%
    \caption{PSNR of rendered scenes using Baseline vs. Clustered. The PSNR is calculated against the original rendered scene. The last column shows how much thread divergence saving we can get}
    \label{tbl:real_scene_rendered_psnr_divergence}
\end{table}

\begin{figure*}
    \centering
    \includegraphics[width=1\linewidth]{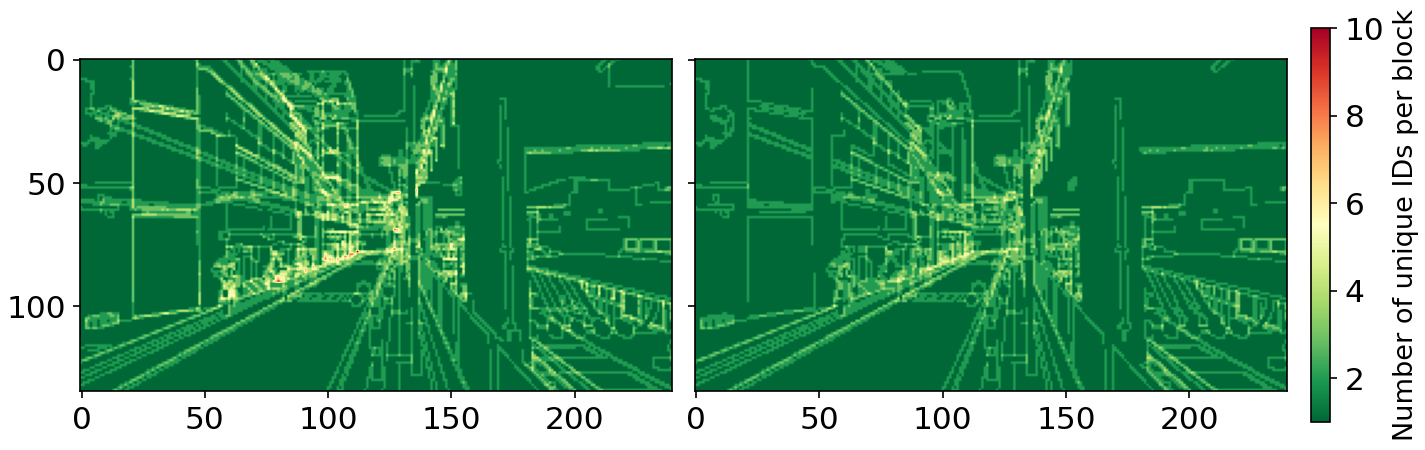}
    
    \includegraphics[width=1\linewidth]{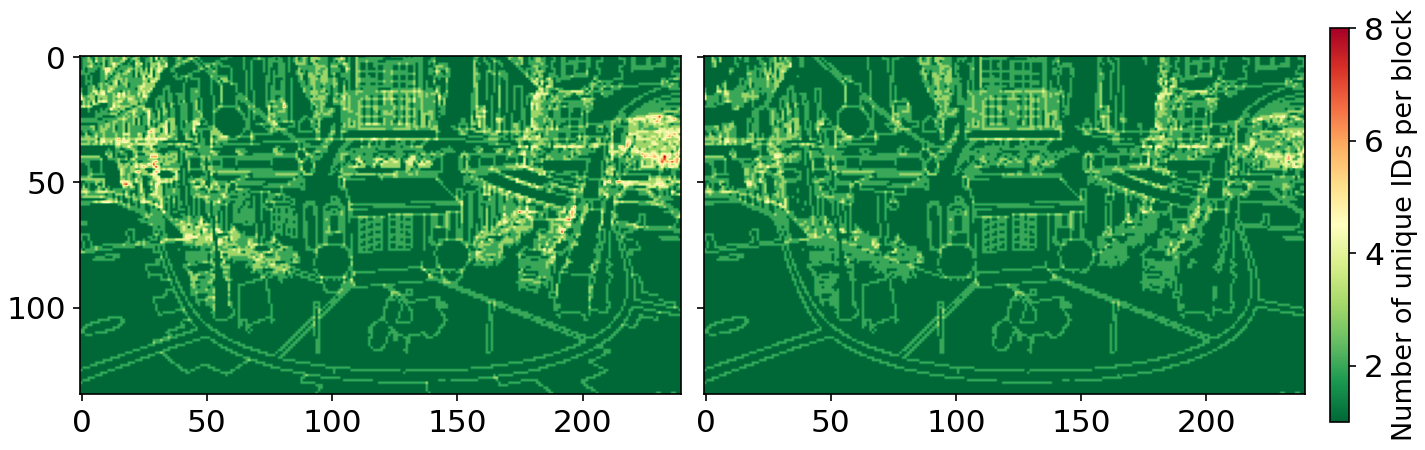}

    \caption{Heatmap of divergence maps for Yokohama (top) and Bistro (bottom). For ToyShop see Figure~\ref{fig:real-scenes-rendered}. The right side of each graph shows the thread divergence in Baseline network, and left side shows how much improved we are getting to reduce the thread divergence using Clustered--green color indicates less divergence. Yokohama, Bistro and ToyShop have the highest (10 and 8) on divergence values as indicated by colorbar.}
    \label{fig:visualize_divergence}
\end{figure*}



\subsection{Evaluation of Thread Divergence}
\begin{figure}
    \centering
    \begin{subfigure}[t]{0.33\linewidth}
        \centering
        \includegraphics[width=\linewidth]{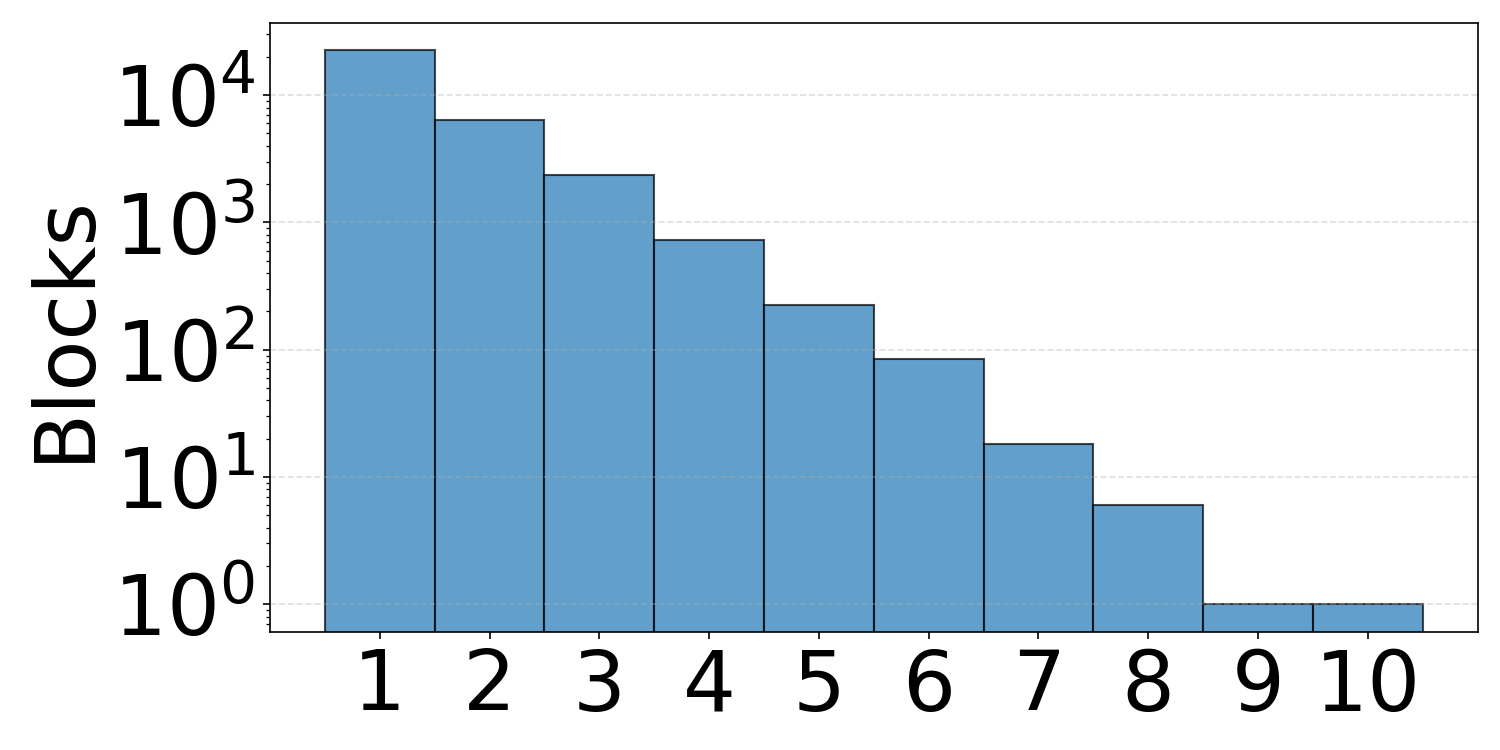}
    \end{subfigure}\hfill
    \begin{subfigure}[t]{0.33\linewidth}
        \centering
        \includegraphics[width=\linewidth]{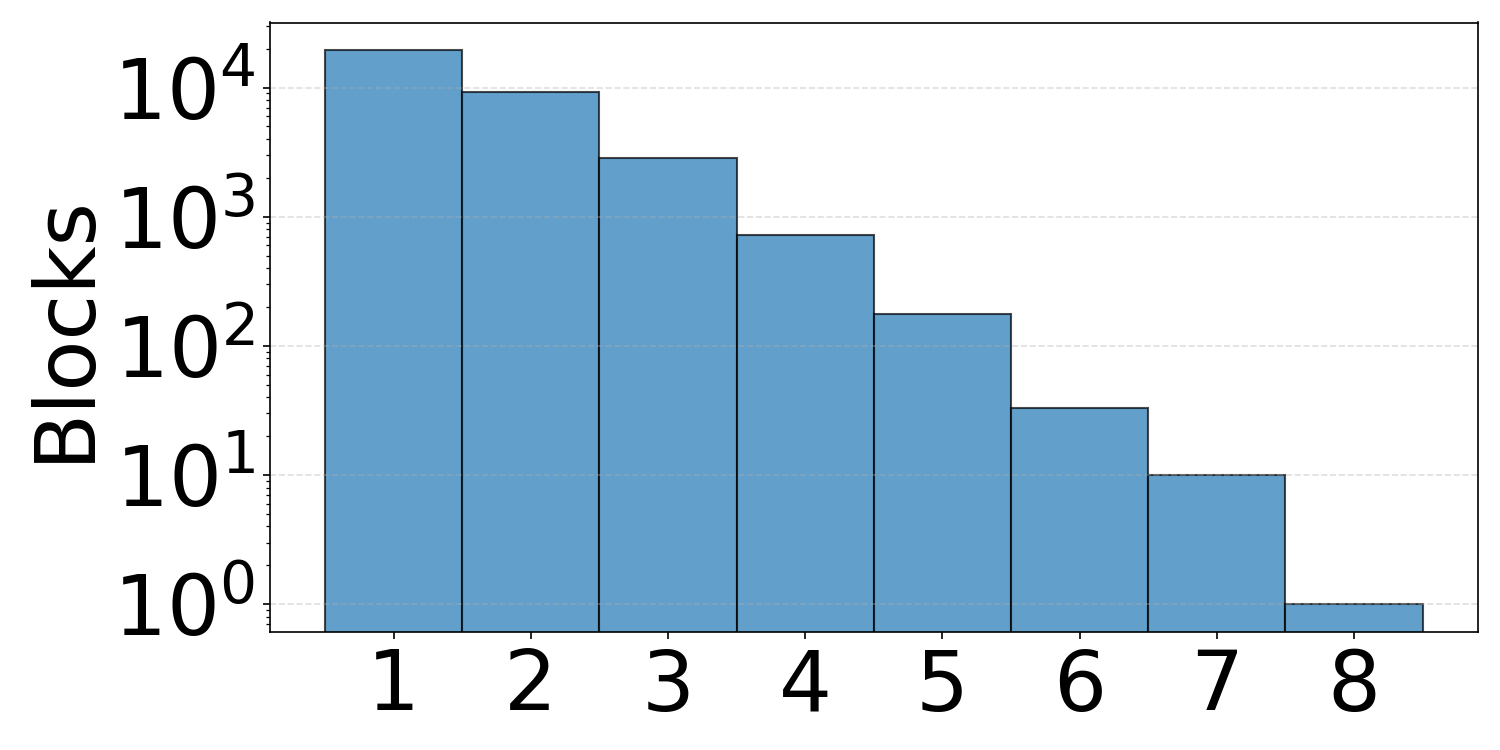}
    \end{subfigure}\hfill
    \begin{subfigure}[t]{0.33\linewidth}
        \centering
        \includegraphics[width=\linewidth]{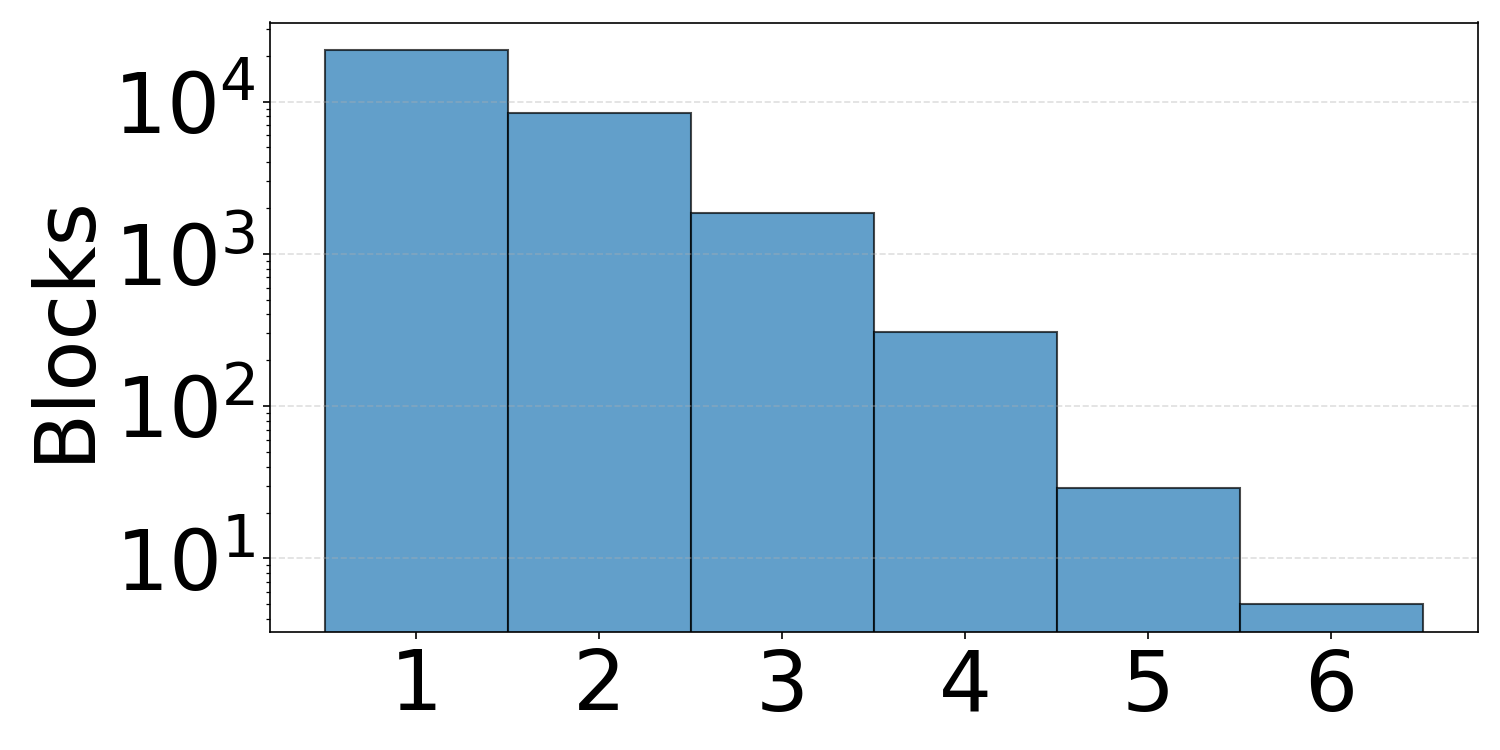}
    \end{subfigure}    

    \begin{subfigure}[t]{0.33\linewidth}
        \centering
        \includegraphics[width=\linewidth]{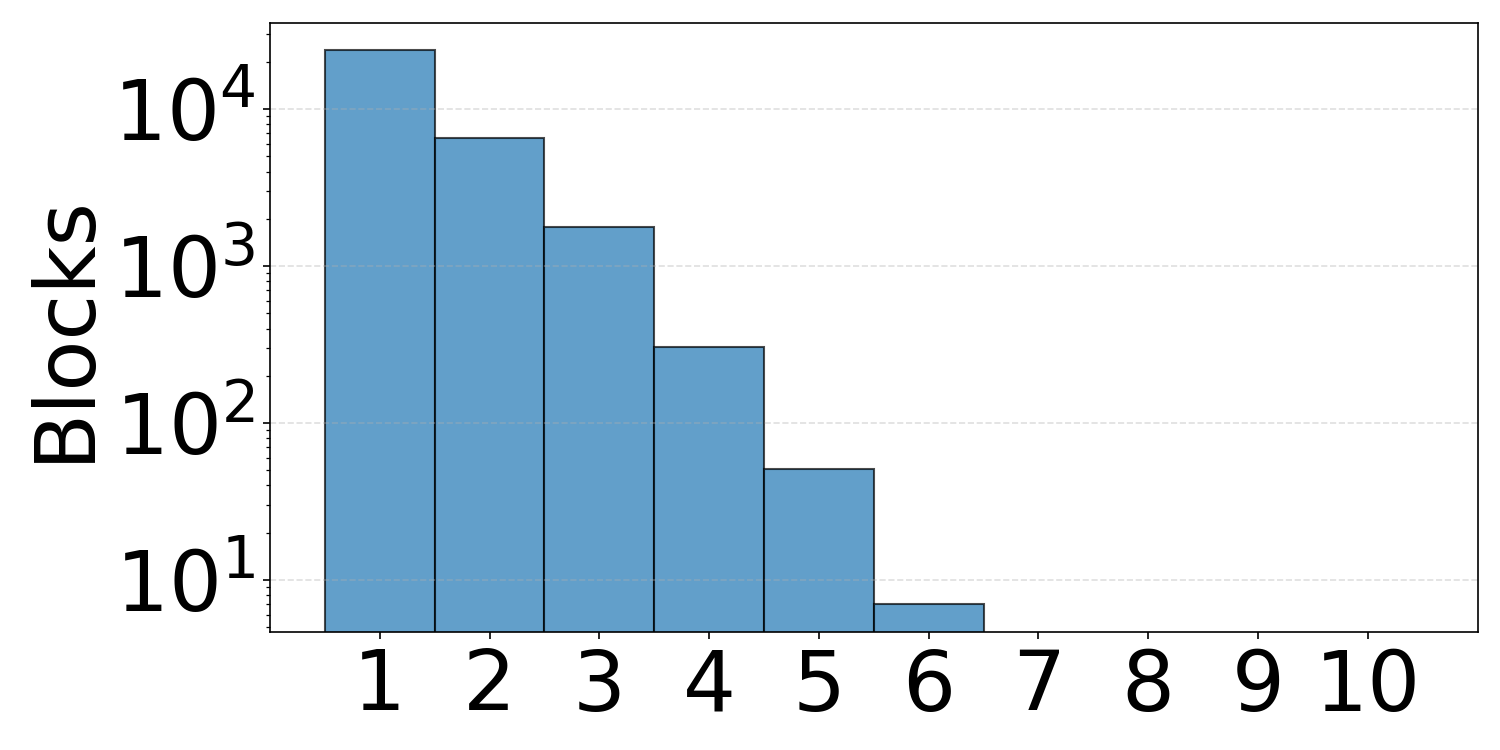}
        \caption{Yokohama}
    \end{subfigure}\hfill
    \begin{subfigure}[t]{0.33\linewidth}
        \centering
        \includegraphics[width=\linewidth]{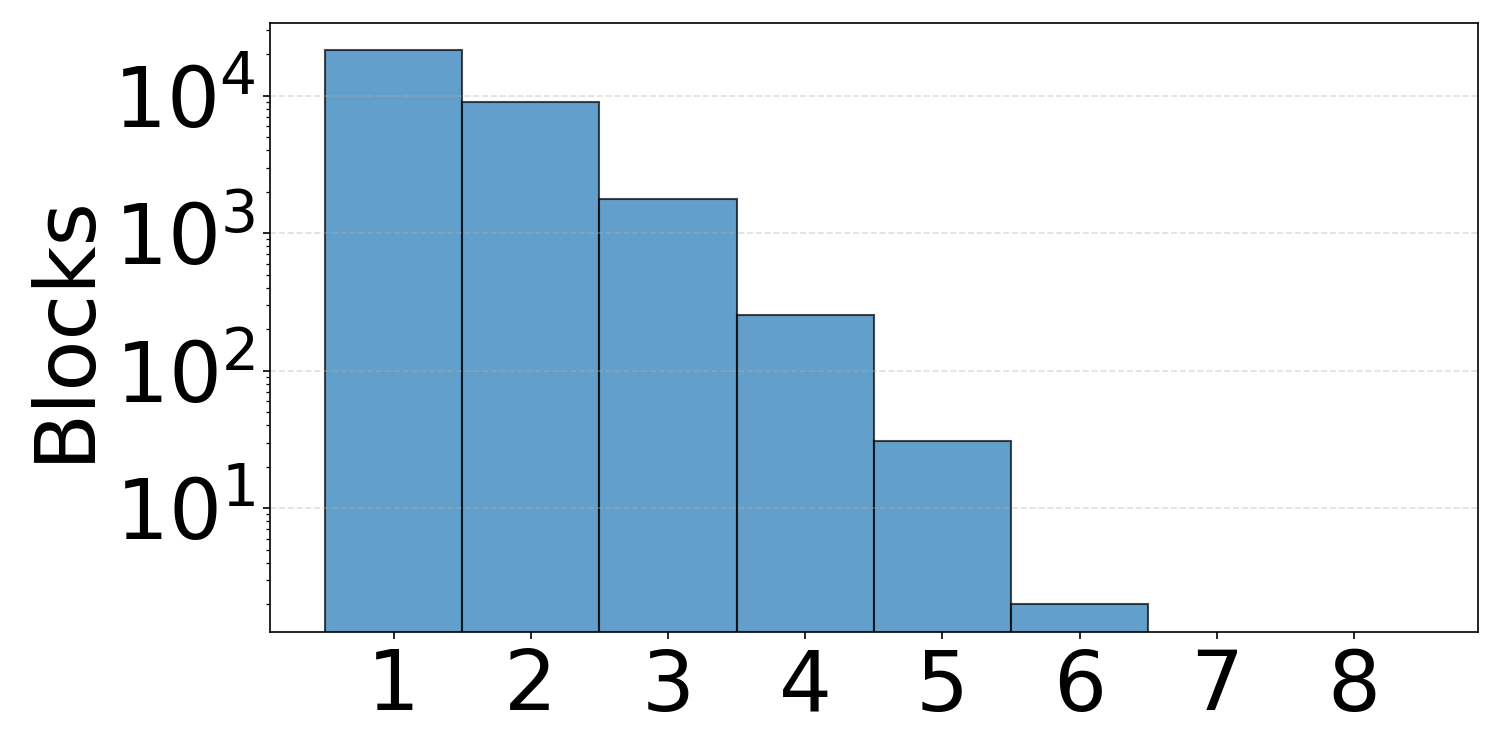}
        \caption{Bistro}
    \end{subfigure}\hfill
    \begin{subfigure}[t]{0.33\linewidth}
        \centering
        \includegraphics[width=\linewidth]{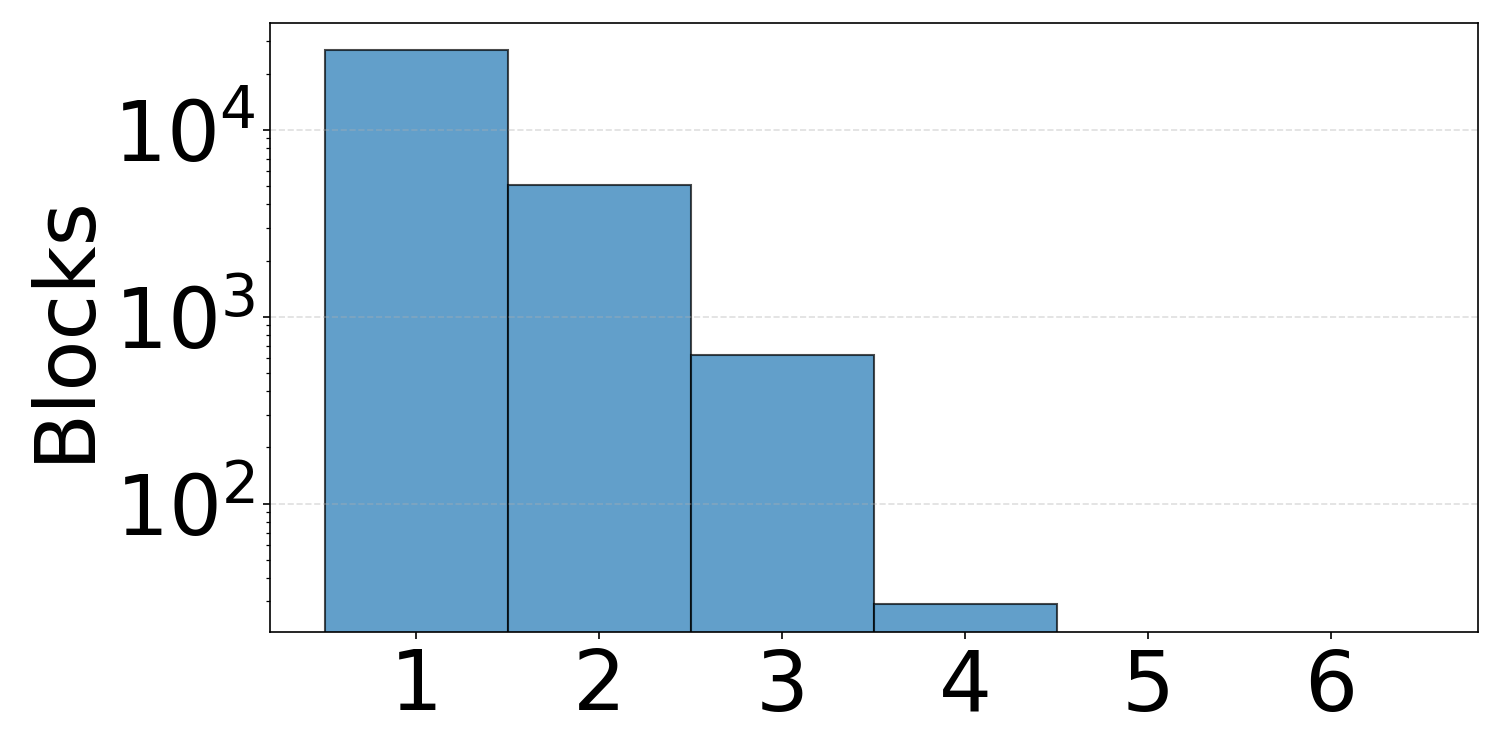}
        \caption{ToyShop}
    \end{subfigure}    
    
    \caption{Divergence map histogram --Baseline vs. Clustered. Top row: Baseline divergence map histogram, and the bottom row Clustered divergence map histogram. The x-axis is the block ids.  The histogram shifts to left compared to top vs. bottom meaning there is less thread divergence in bottom (Clustered). The y-axis is the number of thread blocks that uses the specific texture on x-axis. 
    }
    \label{fig:divergence_histogram}
\end{figure}

We present empirical results demonstrating how our approach reduces thread divergence. The primary objective of this work is to reduce thread divergence in NTC; to that end, we report measurements of thread divergence as defined in Section~\ref{sec:thread_divergence}. In Table~\ref{tbl:real_scene_rendered_psnr_divergence}, the last column lists the number of divergent threads for the Baseline and Clustered approaches. For the three scenes, our approach yields 11{,}222, 13{,}373, and 6{,}407 divergent threads, which correspond to approximately 25\%, 25\%, and 52\% fewer divergent threads than the Baseline, respectively. This reduction in divergence is clearly visible in the divergence-map histograms in Figure~\ref{fig:divergence_histogram} which shows how the divergence distributions shift from right to left for all three scenes, indicating reduced divergence with our approach.
The amount of divergence reduction depends on the scene and on multiple factors, including texture similarity and clustering size. The ToyShop scene exhibits a particularly large reduction in divergence because its textures are less divergent. Our method not only increases the number of non-divergent blocks, but also reduces the amount of divergence within individual blocks, as illustrated in Figure~\ref{fig:visualize_divergence} for Yokohama and Bistro scenes. 


\begin{figure}
    \centering
    \includegraphics[width=0.6\linewidth]{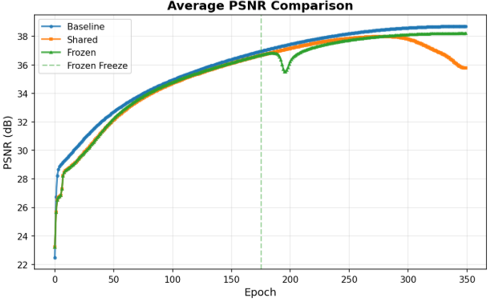}
    \caption{Is freezing decoder helps? Blue is the baseline, Organge is the network without freezing decoder, Green is the decoder freezed at epoch 175. }
    \label{fig:freezing_vs_none_freeze}
\end{figure}

\subsection{Shared Decoder Training}
\label{subsec:shared_decoder_training}
One of the key outcomes of our training experiments is that we freeze the decoder in the shared networks (Cluster, Random, Category). Freezing the decoder improves both stability and accuracy. 

We address two main questions regarding freezing: (1) Does freezing the decoder help? and (2) When should we freeze the decoder? After a comprehensive study across different neural network architectures and multiple texture datasets, we conclude that freezing the decoder leads to higher accuracy compared to not freezing it. As shown in Figure~\ref{fig:freezing_vs_none_freeze}, the accuracy drops at the moment we freeze the decoder, but after a few epochs it surpasses the accuracy of training without freezing. We hypothesize that freezing stabilizes training and helps mitigate overfitting to specific textures, which is particularly important because the decoder is shared across textures. This training receipt helps us to move the burden from decoder to individual encoders and further resulting (as shown in  Figure~\ref{fig:freezing_vs_none_freeze}) recovering the accuracy which is crucial. In subsection~\ref{subsec:ablation_study}, we provide ablation study about how to find the optimal decoder freezing point. 

\subsection{Performance analysis}

We evaluated the performance of our proposed shared decoder approach — Figure~\ref{fig:method_overview} (b) by comparing it against the standard single encoder-single decoder NTC (or Baseline) — Figure~\ref{fig:method_overview} (a), implemented in HLSL on a Radeon\texttrademark~RX 9070 XT GPU. Architecture as in Figure~\ref{fig:method_overview} (c) can do the clustering offline, thus we focus on (a) and (b). Tables \ref{tab:performance_5dec} and~\ref{tab:performance_10dec} present the inference times (in microseconds) for FP16 configurations with 5 and 10 decoders uniformly distributed across a wave, measured across various output image resolutions at two GPU frequencies (1.6 GHz and 2.4 GHz). We compare our Shared-Decoder (baseline) against two non-shared implementations: Non-Shared WMMA (using Wave Matrix Multiply-Accumulate) and Non-Shared FMA (using standard fused multiply-add). Since the MLP is the primary computational bottleneck of the NTC model, the reduced thread divergence achieved through the shared decoder yields substantial speedups, averaging 848\% over Non-Shared WMMA and 795\% over Non-Shared FMA for 10 decoders at 1.6 GHz. The speedup is consistent across image resolutions, and the higher 2.4 GHz GPU frequency reduces absolute inference times while maintaining comparable relative speedups. 

Notably, the Non-Shared WMMA implementation is faster than Non-Shared FMA in the 5-decoder configuration, while showing slightly higher inference times in the 10-decoder configuration, suggesting different scaling characteristics between these two approaches.





\begin{table}[htbp]
\centering
\caption{Performance comparison across different configurations with 5 decoders}
\label{tab:performance_5dec}
\small
\begin{tabular}{@{}lrrr@{\hspace{1em}}rrr@{}}
\toprule
& \multicolumn{3}{c}{GPU Freq: 1.6 GHz} & \multicolumn{3}{c}{GPU Freq: 2.4 GHz} \\
\cmidrule(lr){2-4} \cmidrule(lr){5-7}
Output & Shared- & Non-Shared & Non-Shared & Shared- & Non-Shared & Non-Shared \\
Image & Decoder & WMMA & FMA & MLP & WMMA & FMA \\
\midrule
512²    & 55   & 259   & 394   & 37   & 175   & 270   \\
1024²  & 205  & 1006  & 1530  & 139  & 678   & 1031  \\
2048²  & 808  & 4002  & 6085  & 543  & 2691  & 4083  \\
4096²  & 3204 & 16016 & 24315 & 2150 & 10741 & 16294 \\
8192²  & 12771 & 63851 & 97229 & 8594 & 42937 & 65187 \\
\bottomrule
\multicolumn{7}{@{}l@{}}{\footnotesize Avg. speedup: 1.6 GHz (391.36\%, 647.20\%); 2.4 GHz (391.10\%, 647.95\%)} \\
\end{tabular}
\end{table}

\begin{table}[htbp]
\centering
\caption{Performance comparison across different configurations with 10 decoders}
\label{tab:performance_10dec}
\small
\begin{tabular}{@{}lrrr@{\hspace{1em}}rrr@{}}
\toprule
& \multicolumn{3}{c}{GPU Freq: 1.6 GHz} & \multicolumn{3}{c}{GPU Freq: 2.4 GHz} \\
\cmidrule(lr){2-4} \cmidrule(lr){5-7}
Output & Shared- & Non-Shared & Non-Shared & Shared- & Non-Shared & Non-Shared \\
Image & Decoder & WMMA & FMA & MLP & WMMA & FMA \\
\midrule
512²    & 56    & 503    & 465    & 38    & 340    & 313    \\
1024²   & 207   & 1950   & 1833   & 140   & 1321   & 1244   \\
2048²   & 805   & 7744   & 7357   & 547   & 5235   & 4976   \\
4096²   & 3194  & 31048  & 29493  & 2161  & 20915  & 19930  \\
8192²   & 12838 & 124288 & 118364 & 8615  & 83597  & 79611  \\
\bottomrule
\multicolumn{7}{@{}l@{}}{\footnotesize Avg. speedup: 1.6 GHz (848.49\%, 795.03\%); 2.4 GHz (846.71\%, 793.66\%)} \\
\end{tabular}
\end{table}

\begin{figure*}
    \centering
    \includegraphics[width=1\linewidth]{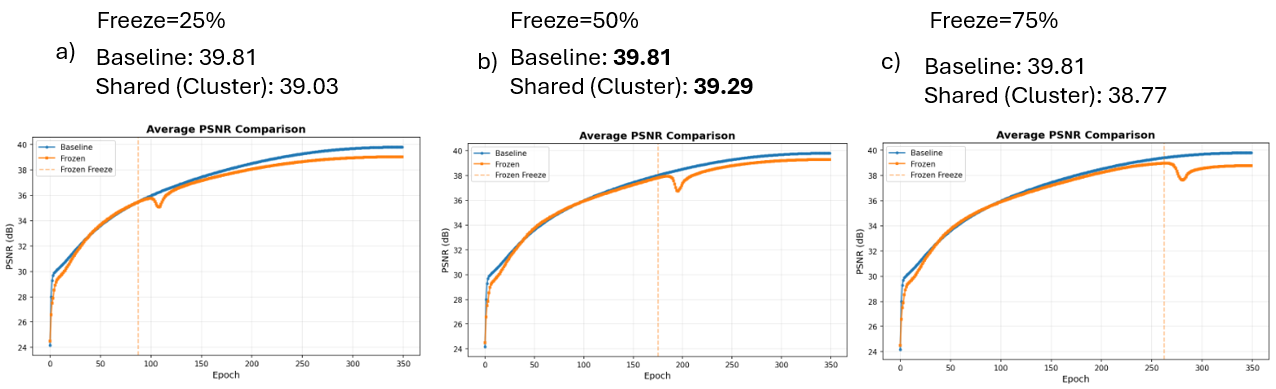}
    \caption{When to freeze? a) Freezing at 25\% of epoch, b) Freezing at 50\% of epoch, c) Freezing at 75\% of epoch. Blue is the baseline, and orange is the Shard network (100 encoders) and textures are randomly selected.}
    \label{fig:freezing_when}
\end{figure*}

\subsection{Ablation Study}
\label{subsec:ablation_study}
In this subsection, first, we study when to freeze the decoder to get better accuracy and stable training, and second, we provide study on the impact of decoder size for shared decoder training. 

\textbf{Freezing Point}: We investigated when to apply freezing and found that freezing around 50\% of the total training epochs usually yields the best accuracy, as shown in Figure~\ref{fig:freezing_when}. In Figure~\ref{fig:freezing_when}, the blue curve is the baseline, and the orange curve is the shared network (100 encoders) with randomly selected textures. As these results show, freezing the decoder at 50\% of the epochs usually produces better performance. Overall, we find that freezing the decoder in our shared networks (Cluster, Random, Category) leads to better generalization, as demonstrated on both the Polyhaven and our rendering results.

\begin{figure}
    \centering
    \includegraphics[width=1\linewidth]{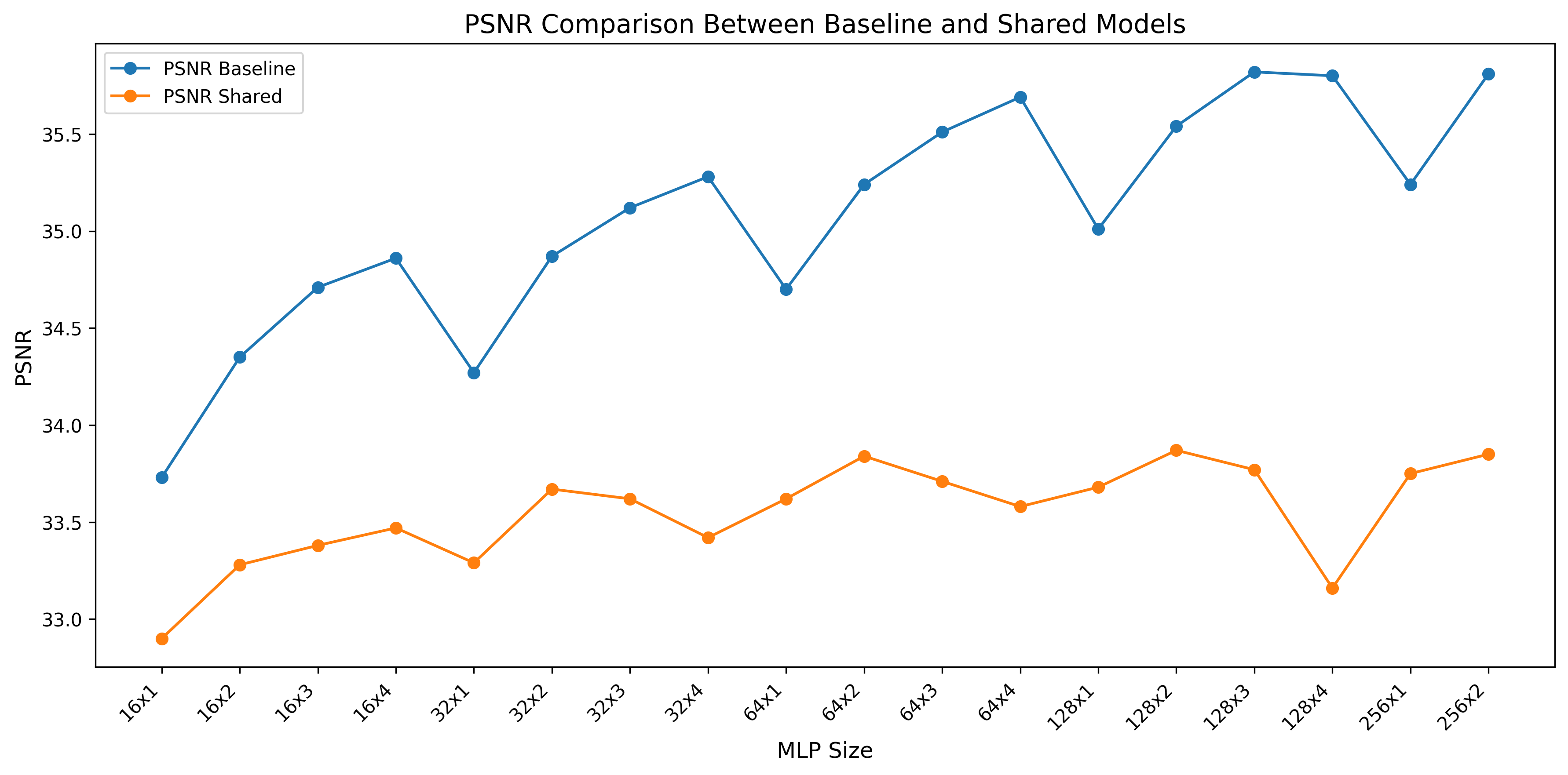}
    \caption{Ablation study of impact of MLP size for baseline vs. shared decoder. The x-axis is the MLP size with format $AxB$ where $A$ is the number of neurons, and $B$ is the number of layers.}
    \label{fig:ml_size_ablation_study}
\end{figure}

\textbf{MLP size impact}: We investigated how the MLP size impacts the accuracy of shard decoder when we increase the depth (number of layers) and breadth (number of neurons) of the MLP. As shown in Figure~\ref{fig:ml_size_ablation_study}, as we increase the MLP size, overall shared decoder accuracy remains comparative with baseline, it gets slightly lower as we increase decoder size significantly. This is due to when we have larger decoder for per texture (baseline), it naturally has higher capacity. Most NTC architectures such as ~\cite{fujieda2024neural, farhadzadeh2024neural, laurent2025hardware, nvidNTC23, ubi24} all have 64 neurons and 2 to 3 layers of MLP. Thus, MLP size of $64x3$ or 64 neurons with $3$ layers are reasonable for most NTC applications -- and anything larger does not have much practical use. 

\section{Discussion and Conclusion}

In this work, we propose a shared decoder architecture along with a clustering-based method to reduce thread divergence in neural texture compression. We comprehensively evaluate the accuracy of our method and the theoretical reduction in thread divergence on both public texture datasets and textures from rendered scenes, and provide a thorough performance and ablation study.

The proposed method significantly reduces thread divergence in NTC, and we highlight several key observations from our results. First, neural networks with a shared decoder can perform well when trained appropriately with proper layer freezing, and their effectiveness depends on the training strategy. Our results indicate that the shared decoder approach not only reduces thread divergence in standalone texture compression, but also in rendering scenarios.

Among the 586 textures and three scenes we evaluated, grouping by similarity yields the best performance, while random grouping is competitive but slightly worse on average. Using a single shared decoder produces the lowest PSNR yet minimizes thread divergence. We view random grouping, clustering, and the single-decoder setup as different points along a spectrum, each offering a different trade-off between accuracy and thread divergence. The single-decoder case can be interpreted as an extreme form of clustering where thread divergence is essentially eliminated. We believe there exists a sweet spot that provides the best balance between the two, and further investigation is needed into how clusters should be formed — for example, based on semantic similarity, low-level features, or cluster size. Regardless of the clustering strategy, our pipeline consistently reduces thread divergence. Our ablation study further confirms that the proposed method consistently accelerates texture decompression across configurations.

\textbf{Limitations and Future Work:} We plan to further investigate the impact of different clustering algorithms on compression accuracy. In this work, we used the OpenAI CLIP model to group textures based on semantic similarity. However, alternative clustering algorithms based on low-level features, such as texture frequency or edge content, are straightforward extensions that we leave for future work. Additionally, the number of clusters $m$ is an important hyperparameter governing the trade-off between divergence reduction and reconstruction quality; a systematic sensitivity analysis of $m$ remains an important avenue for future work. 

\bibliographystyle{ACM-Reference-Format}
\bibliography{sample-base}

\end{document}